\documentclass{article}

\usepackage{PRIMEarxiv}

\usepackage[utf8]{inputenc} 
\usepackage[T1]{fontenc}    
\usepackage{hyperref}       
\usepackage{url}            
\usepackage{booktabs}       
\usepackage{amsfonts}       
\usepackage{nicefrac}       
\usepackage{microtype}      
\usepackage{lipsum}
\usepackage{fancyhdr}       
\usepackage{graphicx}       
\graphicspath{{media/}}     

\usepackage{amsmath,amssymb}
\usepackage{booktabs}
\usepackage{pifont}

\usepackage{xcolor}
\usepackage[normalem]{ulem}

\makeatletter

\newcommand{\etal}{\textit{et al}.\ }
\newcommand{\etc}{\textit{etc}.\ }

\title{Booster-SHOT: Boosting Stacked Homography Transformations for Multiview Pedestrian Detection with Attention}

\author{
  Jinwoo Hwang, Philipp Benz, Tae-hoon Kim \\
  Deeping Source Inc. \\
  Seoul, South Korea \\
  \texttt{\{jinwoo.hwang, philipp.benz, pete.kim\}@deepingsource.io} \\
}

\begin{document}
\maketitle

\begin{abstract}
Improving multi-view aggregation is integral for multi-view pedestrian detection, which aims to obtain a bird's-eye-view pedestrian occupancy map from images captured through a set of calibrated cameras.  Inspired by the success of attention modules for deep neural networks, we first propose a Homography Attention Module (HAM) which is shown to boost the performance of existing end-to-end multiview detection approaches by utilizing a novel channel gate and spatial gate. Additionally, we propose Booster-SHOT, an end-to-end convolutional approach to multiview pedestrian detection incorporating our proposed HAM as well as elements from previous approaches such as view-coherent augmentation or stacked homography transformations. Booster-SHOT achieves 92.9\% and 94.2\% for MODA on Wildtrack and MultiviewX respectively, outperforming the state-of-the-art by 1.4\% on Wildtrack and 0.5\% on MultiviewX, achieving state-of-the-art performance overall for standard evaluation metrics used in multi-view pedestrian detection.
\end{abstract}

\section{Introduction}
\begin{figure}[t]
    \centering
    \includegraphics[width=\linewidth]{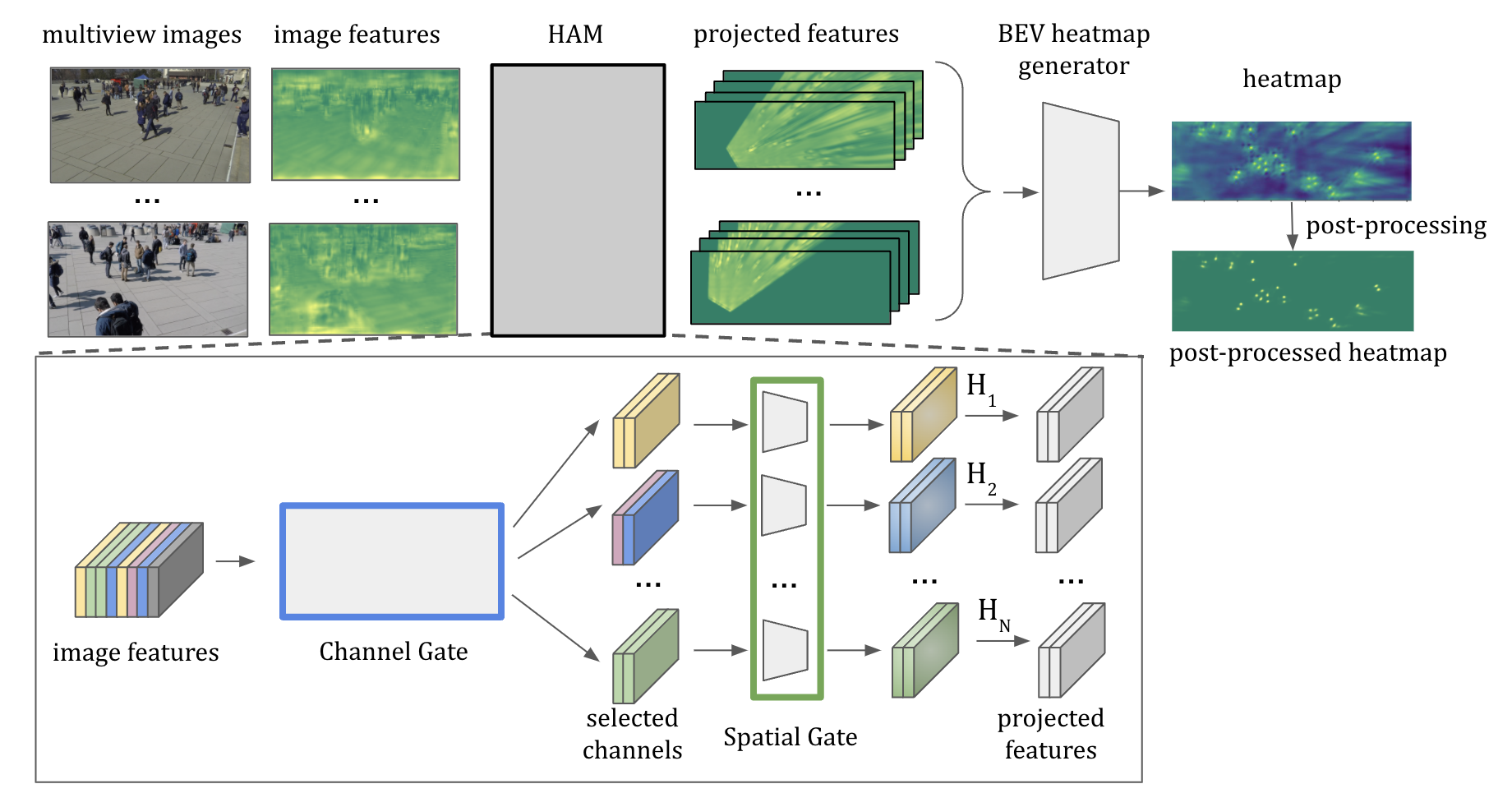}
    \caption{Overview of multiview detection with homography attention module (HAM)}
	\label{fig:multi-view_pedestrian_detection}
\end{figure}

Multi-view detection \cite{sankaranarayanan2008object,aghajan2009multi,hou2020multiview} leverages multiple camera views for object detection using synchronized input images captured from varying view angles. Compared to a single-camera setup, the multi-view setup alleviates the occlusion issue, one of the fundamental problems in many computer vision applications. In this work, we consider the problem of multi-view pedestrian detection. As shown in Figure~\ref{fig:multi-view_pedestrian_detection}, a bird's-eye-view representation is obtained with the synchronized images from multiple calibrated cameras, which is then further used to detect pedestrians in the scene.

A central problem in multi-view detection is to obtain a correct \textit{multi-view aggregation}. The change in viewpoint and occlusions make it challenging to match object features across different view angles. Various works attempted to address this problem, ranging from early approaches leveraging ``classical" computer vision~\cite{alahi2011sparsity}, hybrid approaches further incorporating deep learning, to end-to-end trainable deep learning architectures~\cite{hou2020multiview,hou2021shadow,song2021stacked}.

One core challenge in multiview detection is designing how the multiple views should be aggregated. 
MVDet~\cite{hou2020multiview} proposes a fully convolutional end-to-end trainable solution for the multi-view detection task. MVDet aggregates different views by projecting the convolution feature map via perspective transformation to a single ground plane and concatenating the multiple projected feature maps. Given the aggregated representation, MVDet applies convolutional layers to detect pedestrians in the scene.
Song~\etal~\cite{song2021stacked} identified that the projection of the different camera views to a single ground plane is not accurate due to misalignments. Consequently, they proposed to project the feature maps onto different height levels according to different semantic parts of pedestrians. Additionally, they use a neural-network-based soft-selection module to assign a likelihood to each pixel of the features extracted from the different views. They termed their approach SHOT, due to the use of the Stacked HOmography Transformations.
MVDeTr~\cite{hou2021shadow} extends MVDet by introducing a shadow transformer to attend differently at different positions to deal with various shadow-like distortions as well as a view-coherent data augmentation method, which applies random augmentations while maintaining multiview-consistency. MVDeTr currently constitutes the SotA approach for multiview detection.

In recent years the attention mechanism for deep neural networks has played a crucial role in deep learning~\cite{hu2018squeeze,woo2018cbam,guo2021attention} due to the non-trivial performance gains that it enabled. Attention mechanisms have provided benefits for various vision tasks, e.g. image classification~\cite{hu2018squeeze,woo2018cbam}, object detection~\cite{dai2017deformable,carion2020end}, semantic segmentation~\cite{fu2019dual,yuan2021ocnet}, or Point Cloud Processing~\cite{xie2018attentional,guo2021pct}. However, to this date, no dedicated attention mechanism has been proposed for the task of multiview pedestrian detection.

In this work, we fill this gap and propose an attention mechanism specifically designed to boost existing multiview detection frameworks. Our proposed Homography Attention Module (HAM) is specifically tailored for the core task of multiview aggregation in modern multiview detection frameworks. As shown in the lower part of Figure~\ref{fig:multi-view_pedestrian_detection} our proposed solution consists of a channel gate module and a spatial gate module. The channel gate is directly applied to the accumulated image features from the different views. 
The intuition behind our \textit{channel gate} is that different channels hold meaningful information for different homographies. 
The channel gate is followed by our \textit{spatial gate}. We conjecture, that for each view and homography combination different spatial features are of higher importance. Our proposed attention mechanism can be readily plugged into existing methods.

We also combine insight from previous approaches and HAM to propose Booster-SHOT, a new end-to-end multiview pedestrian detection framework. Our experimental results show that both incorporating HAM into previous frameworks and Booster-SHOT improves over previous multiview detection frameworks and achieves state-of-the-art performance. Additionally, we provide quantitative and qualitative results to verify and justify our design choices.

\section{Related work}
\subsection{Multiview Detection}
It can be difficult to detect pedestrians from a single camera view, due to crowded scenes and occlusions. Hence, many research works study pedestrian detection in the multi-camera setup. Multiple calibrated and synchronized cameras capturing a scene from different view angles can provide a richer representation of the environment. Camera calibrations additionally produces a correspondence between each location on the ground plane and its bounding boxes in multiple camera views. This is possible since a 2D bounding box can be calculated, once an average human height and width is assumed via perspective transformation. 

Research in multiview pedestrian detection has been explored intensively in the past. Early methods mainly relied on background subtraction, geometric constraints, occlusion reasoning, \etc~\cite{fleuret2007multicamera,sankaranarayanan2008object,berclaz2011multiple}.
Given different views from two to four video streams Fleuret~\etal~\cite{fleuret2007multicamera}, first estimates a probabilistic occupancy map of the ground plane via a generative model, which is followed by a tracking mechanism to track up to six individuals. 
The work by Sankaranarayanan~\etal~\cite{sankaranarayanan2008object} emphasizes how geometric constraints in multi-camera problems can be leveraged for detection, tracking, and recognition. 
Similar to~\cite{fleuret2007multicamera}, Coates and Ng~\cite{coates2010multi} also leverage a probabilistic method to fuse the outputs of multiple object detectors from different views to enhance multi-view detection in the context of robotics. 
Using the k-shortest paths algorithm Berclaz~\etal~\cite{berclaz2011multiple} propose a multiple object tracking framework, requiring as input an occupancy map from a detector. Their algorithm handles unknown numbers of objects while filtering out false positives and bridging gaps due to false negatives.
Similar to previous approaches, given the output of several object detectors for different viewpoints Roig~\etal~\cite{roig2011conditional} estimate the ground plane location of the objects. They model the problem via Conditional Random Fields (CRFs) to simultaneously predict the labeling of the entire scene. 

With the success of deep learning, deep neural networks have also been successfully applied to the multi-view detection problem~\cite{chavdarova2017deep,baque2017deep,hou2020multiview,song2021stacked,hou2021shadow}.
Chavdarova and Fleuret~\cite{chavdarova2017deep} propose an end-to-end deep learning architecture that combines the early layers of pedestrian detectors for monocular views with a multiview network optimized for multi-view joint detection.
Baqu{\'e}~\etal~\cite{baque2017deep} identify that the performance of multiview systems degrades significantly in crowded scenes. To overcome this, they propose a hybrid approach of CRFs and CNNs. To perform robustly in crowded scenes their end-to-end trainable model leverages high-order CRF to model occlusions.

Recently, MVDet~\cite{hou2020multiview}, an end-to-end trainable multiview detector has been proposed. To aggregate cues from multiple views, MVDet transforms feature maps obtained from multiple views to a single ground plane. To aggregate information from spatially neighboring locations, MVDet leverages large kernel convolutions on the multiview aggregated feature map. 
MVDet has been further extended through stacked homographies and shadow transformers~\cite{song2021stacked,hou2021shadow}
Song~\etal~\cite{song2021stacked}  identified that the projection to a single view in MVDet~\cite{hou2020multiview} leads to inaccuracies in the alignments. Consequently, they propose to approximate projections in 3D world coordinates via a stack of homographies. Additionally, to 
assemble occupancy information across views, they propose a soft selection module. The soft-selection module predicts a likelihood map that assigns each pixel of the extracted features extracted from individual views to one of the homographies. 
MVDeTr~\cite{hou2021shadow}, adopts shadow transformers to aggregate multiview information, which attend differently based on position and camera differences to deal with shadow-like distortions. Additionally, MVDeTr introduces a view-coherent data augmentation method, which applies random augmentations while maintaining multiview consistency. To the best of our knowledge, MVDeTr currently constitutes the SOTA approach for multiview pedestrian detection. 

\subsection{Attention Mechanism in Computer Vision}
Attention mechanisms for computer vision emphasize more important regions of an image or a feature map and suppress less relevant parts~\cite{guo2021attention}. They can be broadly divided into channel attention, spatial attention, and a combination of the two variants.
\\
\textbf{Channel Attention} selects important channels through an attention mask across the channel domain. 
Pioneered by Hu~\etal~\cite{hu2018squeeze} various works have extended upon the Squeeze-and-Excitation (SE) mechanism module~\cite{gao2019global,lee2019srm,yang2020gated,qin2021fcanet}.
\\
\textbf{Spatial Attention} selects important spatial regions of an image or a feature map. 
Early spatial attention variants are based on recurrent neural networks (RNN) ~\cite{mnih2014recurrent,ba2014multiple}.
In the literature various variants of visual attention-based model can be found~\cite{xu2015show,oktay2018attention}
To achieve transformation invariance while letting CNNs focus on important regions, Spatial Transformer Networks~\cite{jaderberg2015spatial} had been introduced. 
Similar mechanisms have been introduced in deformable convolutions~\cite{dai2017deformable,zhu2019deformable}.
Originating from the field of natural language processing, self-attention mechanisms have been examined for computer vision applications~\cite{wang2018non,carion2020end,dosovitskiy2020image,chen2020generative,zhu2020deformable}.

\textbf{Channel Attention \& Spatial Attention} can also be used in combination.
Residual Attention Networks~\cite{wang2017residual} extend ResNet~\cite{he2016deep} through a channel \& spatial attention mechanism on the feature representations.
A spatial and channel-wise attention mechanism for image captioning has been introduced in~\cite{chen2017sca}.
The Bottleneck Attention Module (BAM)~\cite{park2018bam} and Convolutional Block Attention Module (CBAM)~\cite{woo2018cbam} both infer attention maps along the channel and spatial pathway.
While in the previous two methods the channel and spatial pathways are computed separately, triplet attention~\cite{misra2021rotate} was introduced to account for cross-dimension interaction between the spatial dimensions and channel dimension of the input. 
Channel \& spatial attention has also been applied in the context of segmentation~\cite{roy2018recalibrating,fu2019dual}
Further combinations of channel and spatial attention include self-calibrated convolutions~\cite{liu2020improving}, coordinate attention~\cite{hou2021coordinate} and strip pooling~\cite{hou2020strip}.

\section{Preliminaries}
Let the input images for N camera views be $(I^1,...\ , I^N)$. The respective feature maps obtained from the feature extractor in the initial step of the general framework are denoted as $(F^1,...\ ,F^N)$. The intrinsic, extrinsic parameters of the $i$'th camera are $\mathbf{G}^i \in \mathbb{R}^{3 \times 3}$ and $\mathbf{E}^i = [\mathbf{R}^i | \mathbf{t}^i] \in \mathbb{R}^{3 \times 4}$, respectively, where $\mathbf{R}^i$ is the $3 \times 3$ matrix for rotation in the 3D space and $\mathbf{t}^i$ is the $3 \times 1$ vector representing translation.
Following MVDet~\cite{hou2020multiview}, we quantize the ground plane into grids and define an additional matrix $\mathbf{F}^i \in \mathbb{R}^{3 \times 3}$ that maps world coordinates to the aforementioned grid.
\\
For a pixel in image $I^i$, with pixel coordinates $(u,v)$ and its corresponding position in the 3D space $(X,Y,Z)$, we can write the following equation using the pinhole camera model:
\begin{equation} 
\label{eq:1}
    \begin{pmatrix}
        u \\ v \\ 1
    \end{pmatrix}
    =
    \lambda \mathbf{G}^i [\mathbf{R}^i | \mathbf{t}^i]
    \begin{pmatrix}
        X \\ Y \\ Z \\ 1
    \end{pmatrix}.
\end{equation}
Here, $\lambda$ is a scaling factor that accounts for possible mismatches between image and real 3D space increments.
The above equation can be written as follows for the ground plane ($Z=0$):
\begin{align} \label{eq:2}
    \begin{pmatrix}
        u \\ v \\ 1
    \end{pmatrix}
    & =
    \begin{bmatrix}
        \theta^i_{11} && \theta^i_{12} && \theta^i_{13} && \theta^i_{14}\\
        \theta^i_{21} && \theta^i_{22} && \theta^i_{23} && \theta^i_{24}\\
        \theta^i_{31} && \theta^i_{32} && \theta^i_{33} && \theta^i_{34}\\
    \end{bmatrix}
\end{align}
\begin{align} \label{eq:3}
    \begin{pmatrix}
        X \\ Y \\ 0 \\ 1
    \end{pmatrix}
    & =
    \begin{bmatrix}
        \theta^i_{11} && \theta^i_{12} && \theta^i_{14}\\
        \theta^i_{21} && \theta^i_{22} && \theta^i_{24}\\
        \theta^i_{31} && \theta^i_{32} && \theta^i_{34}\\
    \end{bmatrix}
\end{align}
\begin{align} \label{eq:4}
    \begin{pmatrix}
        X \\ Y \\ 1
    \end{pmatrix}
    & = 
    \mathbf{\Theta}^i
    \begin{pmatrix}
        X \\ Y \\ 1
    \end{pmatrix}.
\end{align}
We can apply the inverse of $\mathbf{\Theta}^i$ to both sides of the equation and multiply by $\mathbf{F}^i$ to obtain a matrix mapping from image coordinates directly to the ground plane grid.
That matrix can be written as 
\begin{equation} \label{eq:5}
    \mathbf{H^i} = \mathbf{F}^i(\mathbf{\Theta}^i)^{-1},
\end{equation}
which is a homography matrix.
\\
To expand this approach to planes other than the ground plane ($Z\neq 0$), we adopt SHOT's~\cite{song2021stacked} method and replace $\mathbf{\Theta}^i$ with 
\begin{equation} \label{eq:6}
    \mathbf{\Theta}^i =
    \begin{bmatrix}
        \theta^i_{11} && \theta^i_{12} && \theta^i_{14} + k \Delta z \theta^i_{13}\\
        \theta^i_{21} && \theta^i_{22} && \theta^i_{24} + k \Delta z \theta^i_{23}\\
        \theta^i_{31} && \theta^i_{32} && \theta^i_{34} + k \Delta z \theta^i_{33}\\
    \end{bmatrix}
\end{equation}
where $\theta^i_{j3}\ (j \in \{1,2,3\})$ are the values omitted in Equation~\ref{eq:3}, $\Delta z$ is the distance between homographies, and $k$ is any non-negative integer lower than the total number of homographies (thus denoting all possible heights for the homography). With this new $\mathbf{\Theta}^i$, we can retain the homography matrix representation shown in Equation~\ref{eq:5}.

\section{Methodology}
\subsection{Previous Multiview Detection Methods}
Before presenting our proposed attention module, we outline the previous multiview detection frameworks in which we have implemented and tested its performance. 
MVDet~\cite{hou2020multiview} presented a multiview detection framework that functions as follows: First, the input images from different viewpoints are passed through a generic feature extractor such as ResNet18 with minor modifications. 
The feature maps are passed through an additional convolutional neural network that detects the head and feet of pedestrians before to aid the network during training. Next, the feature maps are projected to the ground plane via homography transformation and concatenated. Additionally, $x,y$ coordinate maps are concatenated to the stack of transformed feature maps as in CoordConv~\cite{liu2018intriguing}. Finally, this is passed through a CNN to output a bird's-eye-view (BEV) heatmap which is then post-processed via thresholding and non-maximum suppression. Extending upon MVDet, MVDeTr~\cite{hou2021shadow} proposed the use of affine transformations (rotation, translation, sheer, scale, cropping), which are view-coherent augmentations. Additionally, the final CNN to generate the BEV heatmap is replaced with a shadow transformer, with the purpose to handle various distortion patterns during multiview aggregation. MVDeTr further replaces the MSE loss used in MVDet with Focal Loss~\cite{law2018cornernet} coupled with an offset regression loss.
While MVDet and MVDeTr both project the feature maps to the ground plane, SHOT~\cite{song2021stacked} proposes to approximate projections in 3D world coordinates via a stack of homographies. 
To show the efficiency of our homography attention module (HAM), we use the approaches as is without modification to their loss or training configuration, and simply plug in our proposed HAM. As the soft selection module in SHOT is rendered obsolete by our proposed HAM, we remove it when comparing the performance of SHOT with our module with the reported values for SHOT.  
Additionally, based on the advances made in previous works and our proposed attention module we further propose Booster-Shot (see Section~\ref{subsec:Booster-SHOT}). 

\begin{table}[t]
\begin{center}
\caption{Settings for each approach}
\label{table:settings_table}
\scalebox{1.0}{
\begin{tabular}{c|cccc}
\toprule
Method      & Aug. & Loss & BEV gen. & Multi Homogr.\ \\
\midrule
MVDet~\cite{hou2020multiview}      & \ding{55} & MSE & CNN & \ding{55}\\
SHOT~\cite{song2021stacked}        & \ding{55} & MSE & CNN & \checkmark\\
MVDeTr~\cite{hou2021shadow}      & \checkmark & Focal & Transformer & \ding{55}\\
Booster-Shot & \checkmark & Focal & CNN, Transformer & \checkmark\\
\bottomrule
\end{tabular}
}
\end{center}
\end{table}

\subsection{Homography Attention Module}
In this section, we introduce our proposed homography attention module (HAM) for boosting the performance of multi-view pedestrian detection. HAM consists of a channel gate and several spatial gates equal to the number of homographies used. Note, that our attention module is specifically designed for view-aggregation in the context of multiview detection and is hence only applied in the multiview aggregation part. The image feature maps are first passed through the channel gate, then the spatial gate, and then finally through the homography, followed by the BEV heatmap generator.

\begin{figure}[t]
    \centering
    \includegraphics[width=\linewidth]{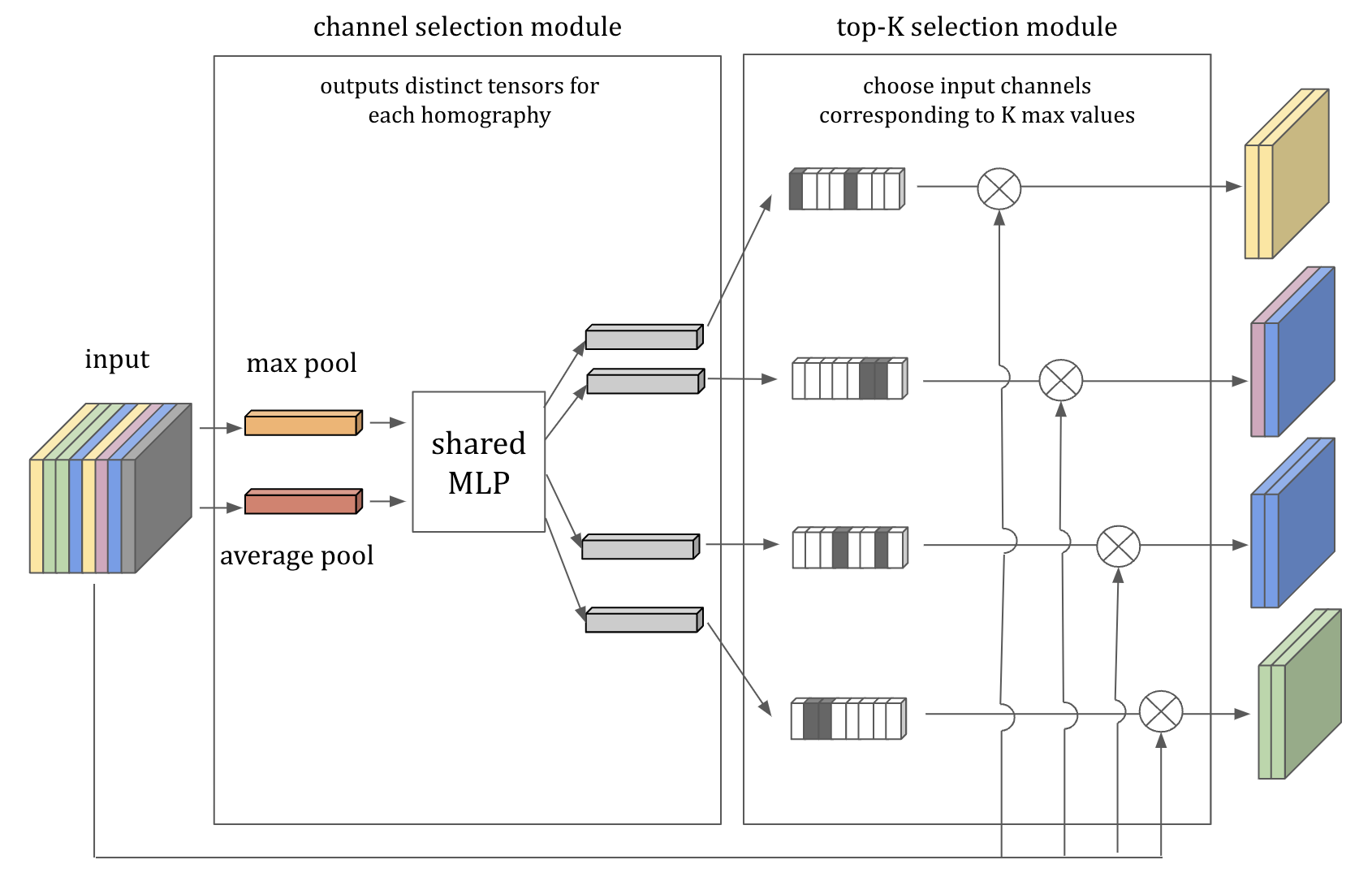}
    \caption{Diagram showing our proposed channel gate.}
	\label{fig:channel_select_module}
\end{figure}

\paragraph{Channel Gate}
Our proposed channel gate follows the intuition that depending on the homography, different channels are of importance. This is in contrast to the approach proposed by SHOT, which feeds all feature maps through each of the homographies. Figure~\ref{fig:channel_select_module} outlines the architecture of our proposed channel gate, which broadly consists of the \textit{channel selection module} and the \textit{top-K selection module}. Given the stack of feature maps acquired from the different views, first the channel selection module is applied. The channel selection module first applies max pooling and average pooling along the spatial dimension. Both pooled feature maps are passed through a shared 2-layer MLP. As the number of channels in the output from the last layer of the MLP is decided by the number of homographies (denoted as $D$) with the number of channels in the input (denoted as $C$), we obtain $C$ channel for each homography, or in other words a $C \times D$ channel output size. Afterward, we apply the softmax function along the channel dimension for each of the outputs. The outputs are then fed into the top-K selection module.
The top-K selection module takes these $D$ different $C$-dimensional outputs and selects the top $K$ largest values. The corresponding top-$K$ selected channels from the original input are then concatenated, resulting in a subset of the original input with $K$ channels. 
In the case of $D=1$ (using only one homography, usually the ground plane), the top-K selection module defaults to an identity function. To retain the channel-wise aspect of our module, in this scenario we multiply the output of the channel selection module element-wise with the input.
This completes the channel gate, which outputs $D$ times $K$-channel feature maps, which are then fed into spatial gate.

\begin{figure}[t]
    \centering
    \includegraphics[width=\linewidth]{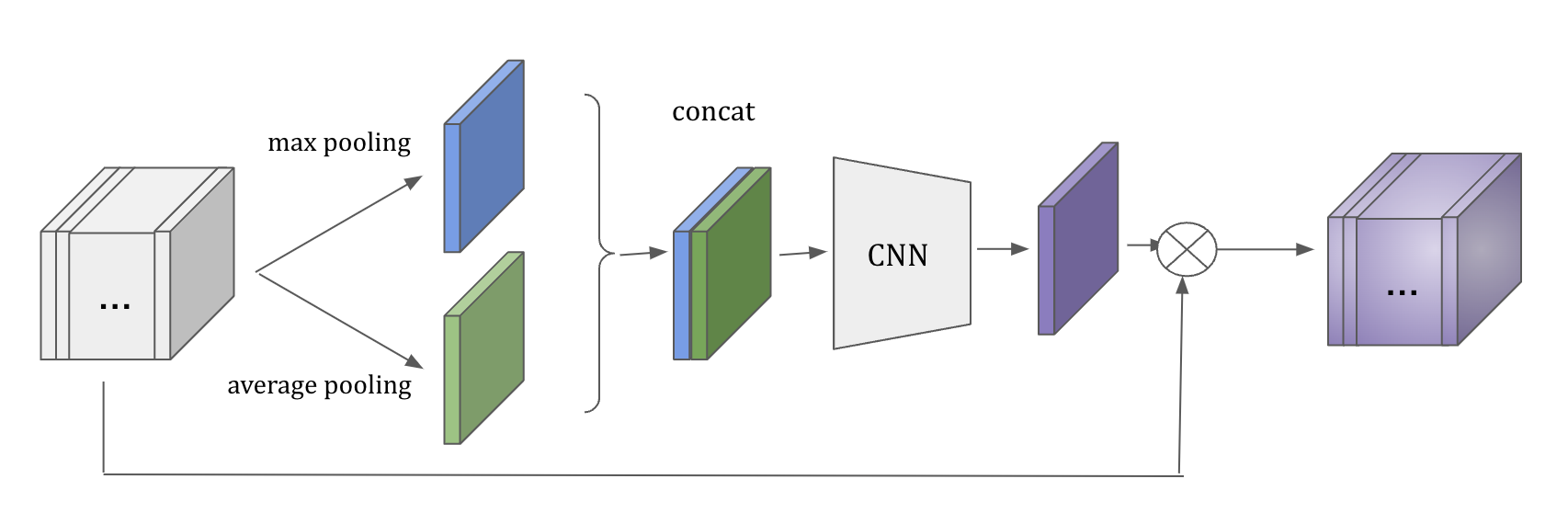}
    \caption{Diagram showing our proposed spatial gate.}
	\label{fig:spatial_attn_module}
\end{figure}

\paragraph{Spatial Gate}
Our spatial gate is motivated by our conjecture that for each view and homography combination different spatial features are of different importance. Figure~\ref{fig:spatial_attn_module} shows the architecture of our spatial gate. The input is max and average pooled along the channel dimension, then concatenated channel-wise. This 2-channel input is then passed through a 2-layer convolutional neural network to generate the spatial attention map. Finally, this spatial attention map is multiplied with the original input element-wise to create an output with dimensions identical to the input. For each homography-path a separate spatial gate is applied. 

\subsection{Booster-SHOT}
\label{subsec:Booster-SHOT}
Given the insights collected from previous approaches in addition to our proposed HAM, we design a multiview pedestrian detection architecture, which we term Booster-SHOT. Booster-SHOT bases itself on SHOT~\cite{song2021stacked}, using their stacked homography approach and leverages MVDeTr's Focal loss and offset regression loss along with the view-coherent augmentation. We retain SHOT's convolutional architecture used to generate the BEV heatmap but remove the soft selection module as the implementation of our module renders it obsolete. Figure~\ref{fig:multi-view_pedestrian_detection} outlines how our proposed module is implemented in Booster-Shot. Table~\ref{table:settings_table} outlines the design choices of Booster-SHOT alongside previous methods.

\section{Experiments}
\subsection{Datasets}
Our method is tested on two datasets for multiview pedestrian detection.
\\
\textbf{Wildtrack}~\cite{chavdarova2018wildtrack} consists of 400 synchronized image pairs from 7 cameras, constituting a total of 2,800 images. The images cover a region with dimensions 12 meters by 36 meters. The ground plane is denoted using a grid of dimensions $480 \times 1440$, such that each grid cell is a 2.5-centimeter by 2.5-centimeter square. Annotations are provided at 2fps and there are, on average, 20 people per frame. Each location within the scene is covered by an average of 3.74 cameras.
\\
\textbf{MultiviewX}~\cite{hou2020multiview} is a synthetic dataset created using human models from PersonX~\cite{sun2019dissecting} and the Unity engine. It consists of $1080 \times 1920$ images taken from 6 cameras that cover a 16-meter by 25-meter area. Per the method adopted in Wildtrack, the ground plane is represented as a $640 \times 1000$ grid of 2.5-centimeter squares. Annotations are provided for 400 frames at 2fps. An average of 4.41 cameras cover each location, while an average of 40 people are present in a single frame.

\subsection{Settings and metrics}
In accordance with the previous methods, we report the four metrics: Multiple Object Detection Accuracy (MODA), Multiple Object Detection Precision (MODP), precision, and recall. 
Let us define $N$ as the number of ground truth pedestrians. 
If the true positives (TP), false positives (FP) and false negatives (FN) are known, precision and recall can be calculated as $\frac{TP}{FP+TP}$ and $\frac{TP}{N}$, respectively. MODA is an accuracy metric for object detection tasks and is therefore obtained by calculating $1-\frac{FP+FN}{N}$. MODP is computed with the formula $\frac{\sum {1-d[d<t]/t}}{TP}$ where $d$ is the distance from a detection to its ground truth (GT) and $t$ is the threshold for a correct detection. We keep the original threshold of 20 that was proposed in SHOT.
Our implementation is based on the released code for MVDet~\cite{hou2020multiview}, SHOT~\cite{song2021stacked}, MVDeTr~\cite{hou2021shadow} and follows the training settings (optimizer, learning rate, \emph{etc}.) for each. For all instances, the input images are resized to $720 \times 1280$ images. The output features are $270 \times 480$ images for MVDet and SHOT and $90 \times 160$ for MVDeTr.
$\Delta z$ (the distance between homographies) is set to $10$ on Wildtrack and $0.1$ on MultiviewX. All experiments are run on two A30 GPUs (depending on the framework) with a batch size of $1$.

For experiments implementing our module in SHOT, our base approach involves selecting the top-32 channels each for 4 homographies. We note that SHOT's base approach uses 5 homographies.

\begin{table}[t]
\caption{Performance comparison (in \%) on Wildtrack and MultiviewX datasets}
    \centering
    \label{table:base_method_performance}
    \scalebox{1.0}{
    \begin{tabular}{c|cccc|cccc}
    \toprule
     & \multicolumn{4}{|c|}{Wildtrack} & \multicolumn{4}{|c}{MultiviewX} \\
    Method & MODA & MODP & precision & recall & MODA & MODP & precision & recall \\
    \midrule
    MVDet & 88.2 & \textbf{75.7} & 94.7 & 93.6 & 83.9 & 79.6 & 96.8 & 86.7 \\
    MVDet + HAM & \textbf{89.4} & \textbf{75.7} & \textbf{95.2} & \textbf{94.1} & \textbf{86.9} & \textbf{81.8} & \textbf{98.6} & \textbf{88.2}\\
    \midrule
    SHOT & 90.2 & 76.5 & 96.1 & 94.0 & 88.3 & 82.0 & 96.6 & 91.5\\
    SHOT + HAM & \textbf{90.5} & \textbf{77.8} & \textbf{96.2} & \textbf{94.2} & \textbf{90.6} & \textbf{82.2} & \textbf{96.8} & \textbf{93.8}\\
    \midrule
    MVDeTr & 91.5 & 82.1 & 97.4 & 94.0 & 93.7 & 91.3 & \textbf{99.5} & 94.2\\
    MVDeTr + HAM & \textbf{92.4} & \textbf{82.9} & 97.2 & \textbf{95.2} & \textbf{94.2} & \textbf{91.4} & \textbf{99.5} & \textbf{94.6}\\
    \midrule
    Booster-Shot + Tr & 92.0 & 82.5 & 96.8 & 95.2 & 94.1 & 91.7 & 98.3 & \textbf{95.7}\\
    Booster-Shot & \textbf{92.9} & \textbf{82.6} & 96.5 & \textbf{96.3} & \textbf{94.2} & \textbf{91.9} & \textbf{99.5} & 94.6\\
    \bottomrule
    \end{tabular}
    }
\end{table}

\subsection{Comparison with previous methods}
As shown above in Table~\ref{table:base_method_performance}, we provide a comparison for the three most recent methods before and after applying our module.
Applying our module to MVDet, SHOT and MVDeTr improved (or matched) all four metrics reported in their respective papers for MultiviewX.
Specifically, the performance of MVDet with our module improves over the reported values for MVDet on MultiviewX by 3.0\%, 2.2\%, 1.8\%, and 1.5\% for MODA, MODP, precision, and recall respectively.
For Wildtrack, the use of our module again improved all four metrics with the exception of MVDeTr. For MVDeTr, our precision was still comparable with the reported value as there was only a 0.2\% decrease in precision while the MODA, MODP, and recall each improved 0.9\%, 0.8\%, and 1.2\% respectively.
When compared with MVDet, SHOT and MVDeTr, Booster-SHOT outperforms them on all metrics except for precision against MVDeTr.

As MVDeTr proposed the shadow transformer as a way to improve performance, we applied it to Booster-SHOT and the results are denoted in Table~\ref{table:base_method_performance} as Booster-SHOT + Tr. However, we were unable to obtain any meaningful improvement over the purely convolutional approach.

\subsection{Ablation Experiments}

\paragraph{Number of homographies}
As shown in SHOT~\cite{song2021stacked}, as using multiple homographies is essentially a quantized version of a 3D projection, using more homographies leads to better performance for multi-view pedestrian detection. As our method assigns fewer channels to each homography as the number of homographies increases, we test the performance of SHOT with our module implemented for 2, 4, 6, and 8 homographies. Overall, all four metrics show improvement as the number of homographies increases (see Table~\ref{table:SHOT_number_of_homographies}). The 6 homography case has the highest MODP and recall while the 8 homography case has the highest precision. Both cases mentioned above have the highest MODA. As the overall performance is very similar, we conclude that the improvement from the increased number of homographies has reached an equilibrium with the decreased number of channels passed to each homography and caused our approach to saturate.

\begin{table}[!htbp]
\begin{center}
\caption{Performance depending on the number of homographies}
\label{table:SHOT_number_of_homographies}
\scalebox{1.0}{
\begin{tabular}{c|c|cccc}
\toprule
& & \multicolumn{4}{c}{MultiviewX} \\
Method & \#H & MODA & MODP & precision & recall \\
\midrule
SHOT       & 5 & 88.3 & 82.0 & 96.6 & 91.5 \\
SHOT + HAM & 2 & 89.4 & 80.8 & 95.2 & 94.2 \\
SHOT + HAM & 4 & 90.6 & 82.2 & 96.8 & 93.8 \\
SHOT + HAM & 6 & \textbf{91.4} & \textbf{83.1} & 97.4 & \textbf{93.9} \\
SHOT + HAM & 8 & \textbf{91.4} & 82.6 & \textbf{97.5} & 93.8 \\
\bottomrule
\end{tabular}
}
\end{center}
\end{table}

\paragraph{Number of top-$K$ channels} 
Our approach initially determined the number of channels selected per homography based on the number of homographies and the number of input channels. For example, our base approach for 128 input channels and 4 homographies involves selecting the top-32 channels for each homography. 
We further test the performance of our module when we fix the number of channels selected per homography (hereon denoted as $K$ in accordance with the name top-K selection) and change the number of output channels accordingly. Setting $K=64$ for 4 homographies and 128 input channels indicates we take the top-64 channels for each homography and output $64 \times 4 = 256$ channels. Table~\ref{table:SHOT_number_of_channels} outlines the results we get for $K=4,8,16,32,64,128$. For MODA, MODP and precision, using the top-16 channels for each homography outperforms the other instances with considerable margins. The top-32 instance (our base approach) improves on the top-16 instance only for recall. We conclude that our channel selection approach is effective in removing irrelevant channels and concentrating relevant information into selected channels for each homography.

\begin{table}[!htbp]
\begin{center}
\caption{Performance depending on the number of selected channels}
\label{table:SHOT_number_of_channels}
\scalebox{1.0}{
\begin{tabular}{c|c|cccc}
\toprule
& & \multicolumn{4}{c}{MultiviewX}\\
Method & $K$ & MODA & MODP & precision & recall\\
\midrule
SHOT + HAM & 4   & 90.6 & 81.8 & 97.7 & 92.7\\
SHOT + HAM & 8   & 90.4 & 82.2 & 97.9 & 92.4\\
SHOT + HAM & 16  & \textbf{91.8} & \textbf{82.6} & \textbf{98.9} & 92.9\\
SHOT + HAM & 32  & 90.6 & 82.2 & 96.8 & \textbf{93.8}\\
SHOT + HAM & 64  & 90.2 & 82.2 & 96.9 & 93.2\\
SHOT + HAM & 128 & 89.2 & 81.8 & 96.0 & 93.0\\
\bottomrule
\hline
\end{tabular}
}
\end{center}
\end{table}

\paragraph{Attention Mechanisms}
In Table~\ref{table:attention_mvdet}, we outline the effects of the channel gate and the spatial gate on MVDet, as well as their combination (HAM). 
It can be observed that both the channel gate and the spatial gate individually improve the performance over MVDet. However, using the channel gate and spatial gate subsequently, in other words HAM, improves in MODA and recall while retaining similar precision compared to MVDet, leading to an overall improvement in performance.

\begin{table}[!htbp]
\caption{Performance of attention modules on MVDet}
\label{table:attention_mvdet}
\centering
\scalebox{1.0}{
\begin{tabular}{c|cccc}
\toprule
& \multicolumn{4}{c}{Wildtrack} \\
Method & MODA & MODP & precision & recall \\
\midrule
MVDet & 88.2 & 75.7 & 94.7 & 93.6 \\
MVDet + Channel Gate & 88.8 & 76.0 & 95.1 & 93.6 \\
MVDet + Spatial Gate & 88.6 & \textbf{76.6} & \textbf{95.5} & 93.0 \\
MVDet + HAM & \textbf{89.4} & 75.7 & 95.2 & \textbf{94.1} \\
\bottomrule
\end{tabular}
}
\end{table}

\subsection{Analysis}

\begin{figure}[!htbp]
    \centering
    \includegraphics[width=0.6\linewidth]{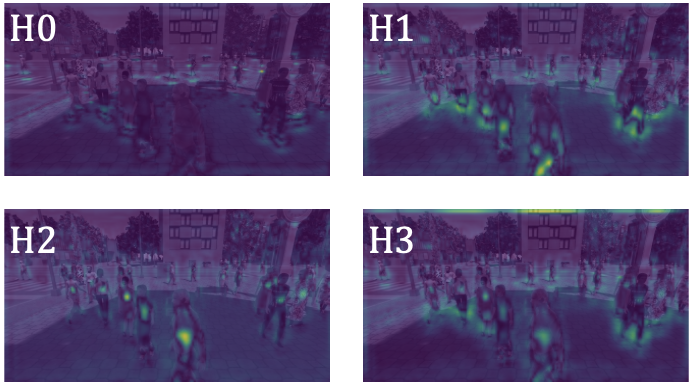}
    \caption{Homography-wise output from channel selection module}
	\label{fig:channel_select_homography}
\end{figure}

\paragraph{Attention for different homographies}
We previously conjectured that the significance of each channel is different for each homography. In the following we validate this hypothesis through empirical evidence.
Figure~\ref{fig:channel_select_homography} shows images created from camera view 1 of the MultiviewX dataset and containing output from the channel selection module corresponding to each homography. The channel selection module output is average pooled channel-wise (in this instance, the output for each homography contains 32 channels) and superimposed onto a grayscale version of the original image from the MultiviewX dataset. Yellow areas indicate high values in the output, indicating that the network is attending strongly to those regions. We denote the ground plane as H0 (homography 0) and number the remaining homographies accordingly. We can observe that the output from the channel selection module is homography-dependent as the yellow areas in all four images differ. We also note that the body parts with the brightest colors align with the height of the homographies. H0 highlights the feet while H1 highlights the lower body, especially around the knee area. H2 and H3 both highlight the upper body but H3 extends a bit farther upwards compared to H2. A similar phenomenon has been reported by the SHOT authors for their soft selection module. However, our channel selection module output shows more distinct highlighting of the body parts and is obtained through a completely different method. Overall, these results support the importance of selecting different channels for different homographies.

\begin{figure}[!htbp]
    \centering
    \includegraphics[width=0.6\linewidth]{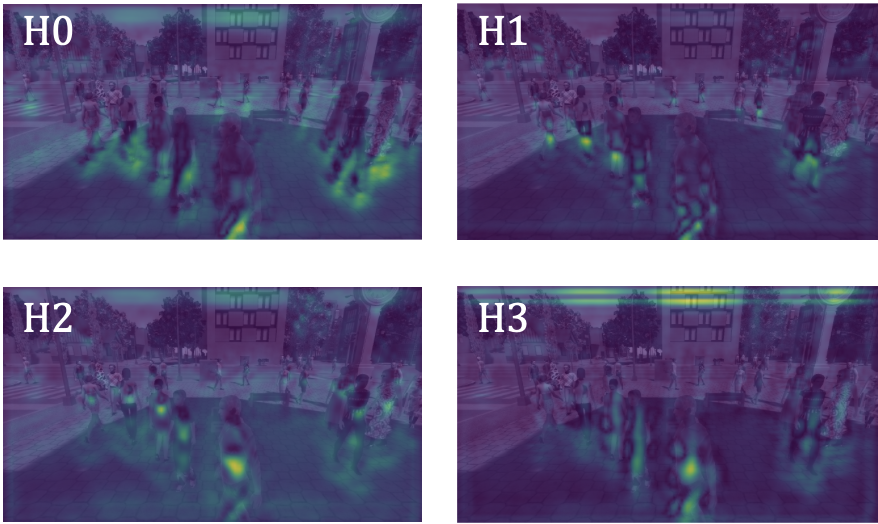}
    \caption{Homography-wise spatial attention maps.}
	\label{fig:spatial_attn_homography}
\end{figure}

Figure~\ref{fig:spatial_attn_homography} shows the attention values from the spatial attention block at the end of our proposed module.
All four attention maps show starkly different distributions, confirming our conjecture that different pixels in the feature map can differ in importance for each homography.

\begin{figure}[!htbp]
    \centering
    \includegraphics[width=0.9\linewidth]{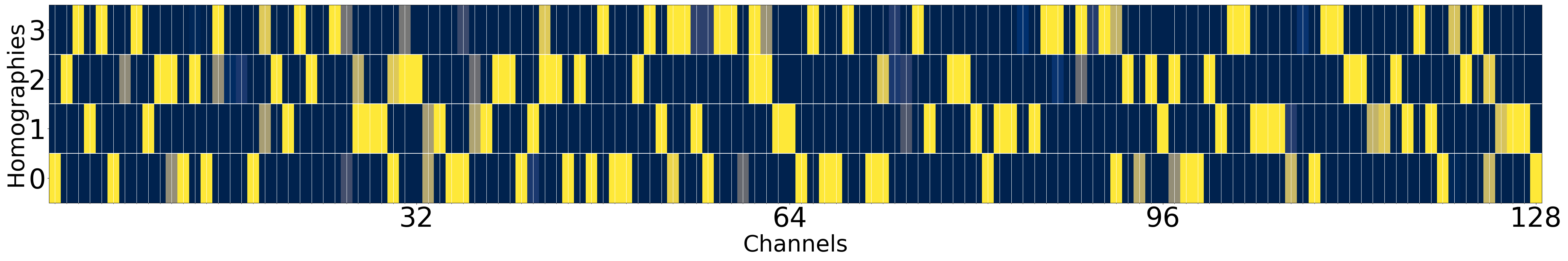}
    \caption{Heatmap representation of channel selection homography-wise. Deeper yellow colors indicates that the channel is selected most of the time while deeper blue colors are assigned to channels that are seldom selected.}
	\label{fig:channel_select_heatmap}
\end{figure}

The results shown above were obtained through an experiment where the distance between homography planes was increased from 10cm to 60cm for MultiviewX. We noticed that, due to the low height of even the top homography plane in the 10cm case (30cm off the ground), the difference between the attention module outputs was not easily noticeable. By increasing the distance between homography planes, we were able to obtain images that clearly show homographies that are higher off the ground attend to higher regions of the human body. 
In addition, we noticed that the foot regression auxiliary loss caused bias toward the foot region in the extracted image features, thus distorting our heatmap visualization of the attention module outputs. As such, the experiments from which Figure~\ref{fig:channel_select_homography}, Figure~\ref{fig:spatial_attn_homography}, Figure~\ref{fig:channel_select_heatmap} and Figure~\ref{fig:channel_select_view} were obtained did not include auxiliary losses during training (details in Appendix~\ref{app:E}).

We further provide results averaged over the entire MultiviewX test dataset. Specifically, we visualize how often certain channels are selected for each homography for a given view. We select Booster-SHOT for this experiment. For each channel, we count the number of times it is selected for each homography and divide by the total number of image pairs in the test set and display the resulting heatmap in Figure~\ref{fig:channel_select_heatmap}. 
First, it can be observed that the channels that are selected often (yellow hues) show almost no overlap across homographies, again providing evidence to our previous claim that different channels attend to different homographies. Although there are minor differences in the specific number of times some channels are chosen, the channels that are selected for the majority of the test set for each homography are unchanged (see Appendix~\ref{app:F}). Interestingly, we also observe that some channels are not selected at all by any homography while other channels appear to be selected by multiple homographies.

\paragraph{Attention across different views}
Figure~\ref{fig:channel_select_view} further presents evidence that our channel selection module output is only homography-dependent. We denote the camera views as C1 (camera 1) through C6 for the homography to the ground plane (H0). For all 6 images, the feet and surrounding area of the pedestrians are highlighted. Therefore, we conclude that the output from the channel selection module attends consistently to certain features across all camera views.

\begin{figure}[!htbp]
    \centering
    \includegraphics[width=0.8\linewidth]{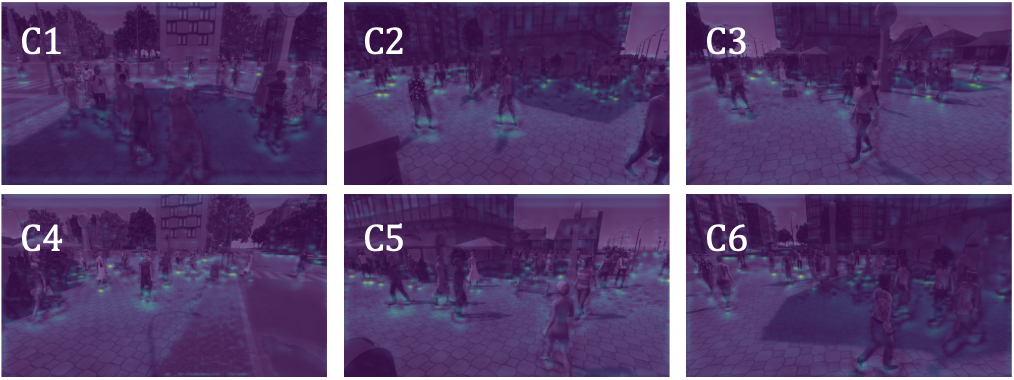}
    \caption{Camera view-wise output from channel selection module}
	\label{fig:channel_select_view}
\end{figure}

\textbf{Generalization across camera views.} 
To evaluate the generalization capabilities for unseen views, we compared different methods on non-overlapping camera subsets from MultiviewX. Adding the HAM module to MVDet and SHOT showed significant improvement in MODA and precision while retaining comparable performance for MODP and recall, providing evidence that HAM also facilitates generalization.
This and further extended analysis results can be found in the Appendix.

\section{Conclusion}
In this work, we propose a homography attention module (HAM) as a way to improve across all existing multiview pedestrian detection approaches. HAM consists of a novel channel gate module that selects the most important channels for each homography and a spatial gate module that applies spatial attention for each homography. In addition, we outline an end-to-end multiview pedestrian detection framework (Booster-SHOT) taking insight from previous approaches while also incorporating our proposed module. For both Booster-SHOT and previous approaches with HAM, we report new state-of-the-art performance on standard benchmarks while providing extensive empirical evidence that our conjectures and design choices are logically sound.


\bibliographystyle{unsrt}  
\bibliography{bibliography}

\appendix
\section{Appendix}

\subsection{Camera-split evaluation}
\label{app:camera-split}
Due to the specific nature of camera pose and environmental factors, multiview pedestrian detection methods are susceptible to overfitting to a single scene. To test generalization capabilities, we follow the experiment in Table 4 in SHOT~\cite{song2021stacked} and train models on a split of cameras of MultiviewX and evaluate the remaining cameras. Specifically, the top-view grid of the ground plane is divided into two separate grids of equal area. As MultiviewX has 6 camera views, 3 cameras cover each of the newly formed grids. We compare our approach to SHOT and MVDet.

The results presented in Table~\ref{tab:split-eval} show that introducing HAM results in significant improvement in MODA and precision for MVDet and SHOT, with MVDet's MODA increasing 27.4\% and SHOT's MODA increasing 14.7\%. Although a similar increase in MODP and recall was not observed, both metrics remain comparable to MVDet and SHOT. Through these results, HAM is shown to boost the generalization ability of both a single-homography (MVDet) approach and a multi-homography (SHOT) approach.

\begin{table}[!htbp]
    \caption{Results from train-test camera split scenario on MultiviewX}
    \label{tab:split-eval}
    \centering
    \scalebox{1.0}{
    \begin{tabular}{c|ccccccccc}
    \toprule
    & MODA & MODP & precision & recall \\
    \midrule
    MVDet               & 33.0 & \textbf{76.5} & 64.5 & \textbf{73.4} \\
    MVDet + HAM         & \textbf{60.4} & 75.2 & \textbf{85.6} & 72.7 \\
    \midrule
    SHOT                & 49.1 & \textbf{77.0} & 73.3 & \textbf{77.1} \\
    SHOT + HAM.         & \textbf{63.8} & 76.6 & \textbf{86.0} & 76.2 \\
    \bottomrule
    \end{tabular}
    }
\end{table}

\subsection{Efficacy of HAM in comparison to existing methods}
We emphasize that the novelty of HAM lies in the architectural integration of the attention mechanism for the specific purpose of multi-view aggregation, for which, to the best of our knowledge, our work is the first. Previous attention mechanisms (e.g. CBAM~\cite{woo2018cbam}, CCG~\cite{abati2020conditional}) are applied at the convolutional blocks in the backbone network, while HAM is applied after the backbone network since it is tailored toward multi-view aggregation. Consequently, HAM can be seen as complementary to existing attention mechanisms.

To illustrate the importance of the design choices of HAM we compare it with the naive integration of CBAM and CCG into Booster-SHOT on MultiviewX. In the two latter implementations, CBAM and CCG come after the feature extractor in place of HAM. To provide a common baseline for HAM, CBAM, and CCG, we provide additional results for "BoosterSHOT without attention". This implementation is equivalent to SHOT~\cite{song2021stacked} with Focal Loss and training-time augmentations.

As shown in Table~\ref{table:attention-modules}, BoosterSHOT is shown to outperform all of the compared methods across the board. Only BoosterSHOT without attention shows better results in precision, a very saturated metric for which BoosterSHOT shows only 0.1\% lower performance. In addition, when compared with BoosterSHOT without attention, adding CBAM and CCG showed a 0.1\% and 0.2\% increase in MODA while adding HAM boosted MODA by 1.6\%.

\begin{table}[t]
\begin{center}
    \caption{BoosterSHOT performance with HAM vs pre-existing attention mechanisms}
    \label{table:attention-modules}
    \scalebox{1.0}{
    \begin{tabular}{c|ccccccccc}
    \toprule
    & MODA & MODP & precision & recall \\
    \midrule
    Booster-SHOT w/o attention  & 92.9 & 91.1 & \textbf{99.4} & 93.4 \\
    \midrule
    Booster-SHOT (CBAM)   & 93.0 & 90.7	& 97.7 & 95.2 \\
    Booster-SHOT (CCG)    & 93.1 & 91.4 & 98.9 & 94.1 \\
    Booster-SHOT          & \textbf{94.5} & \textbf{92.0} & 99.3 & \textbf{95.2} \\
    \bottomrule
    \end{tabular}
    }
\end{center}
\end{table}

\subsection{BoosterSHOT performance under various settings for distance between homography planes}
\label{app:C}
In the main text, we visualized results for BoosterSHOT with 60cm between each homography plane. To explore whether the distance between homography planes has any significant impact on the performance of BoosterSHOT, we performed additional experiments with the distance between homography planes set to 20cm and 40cm, respectively. As shown in Table~\ref{table:homography-distance}, variations of BoosterSHOT showed comparable performance. Variations with more distance between homography planes even showed a consistent 0.3\$ increase in MODA while other metrics remained comparable to or better than MVDeTr.

\subsection{Spatial gate results under various settings for distance between homography planes}
\label{app:D}

\begin{table}[t]
\begin{center}
    \caption{BoosterSHOT performance based on distance between homography planes}
    \label{table:homography-distance}
    \scalebox{1.0}{
    \begin{tabular}{c|ccccccccc}
    \toprule
    & MODA & MODP & precision & recall \\
    \midrule
    Booster-SHOT ($\Delta z$=$0.2$) & \textbf{94.4} & 92.4 & 99.0 & 95.3 \\
    Booster-SHOT ($\Delta z$=$0.4$)  & \textbf{94.4} & 92.4 & 99.0 & 95.3 \\
    Booster-SHOT ($\Delta z$=$0.6$) & \textbf{94.4} & \textbf{92.5} & 98.2 & \textbf{96.2} \\
    Booster-SHOT          & 94.1 & 91.7 & 98.3 & 95.7 \\
    \midrule
    MVDeTr          & 93.7 & 91.3 & \textbf{99.5} & 94.2 \\
    \bottomrule
    \end{tabular}
    }
\end{center}
\end{table}

\begin{figure}[!htbp]
    \centering
    \includegraphics[width=0.8\linewidth]{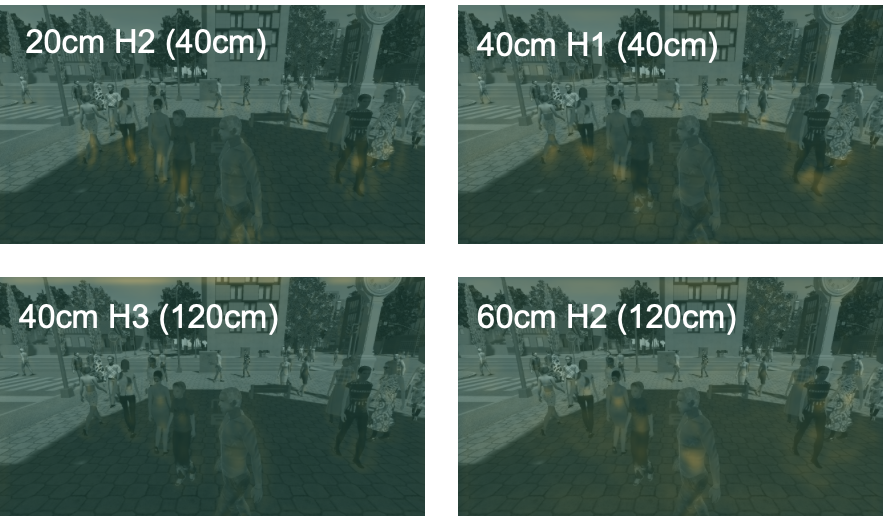}
    \caption{Comparison of spatial attention maps for homography planes of equal height}
	\label{fig:homography_same_height}
\end{figure}

We further investigate the relation between our spatial attention heatmaps and the distance between homography planes. To this end, we provide Figure~\ref{fig:homography_same_height} showing 4 spatial attention heatmaps. The first row shows spatial attention heatmaps for a homography plane at 40cm above the ground and the second row shows heatmaps for a homography plane at 120cm above the ground. Heatmaps corresponding to the same height attend to similar regions of the image, while those at different heights show different patterns when it comes to the highlighted regions. This is an indication of the validity of our claim that spatial attention is reliant on and consistent with the height of each of the homography planes.

\subsection{Effect of per-view loss on model performance}
\label{app:E}
To quantify the effect of the per-view loss on BoosterSHOT, we experimented with 3 different settings. Per-view loss indicates any loss used only during training, such as losses for foot position regression in each camera view. We show the results for BoosterSHOT with no per-view losses, with a loss term for foot position regression, and with loss terms for both foot and head position regression in Table~\ref{table:per_view_loss}. For the pre-existing literature, MVDet used both head and foot loss, while SHOT and MVDeTr used only foot regression as a subtask during training. Introducing a head position regression subtask during the training phase shows a non-negligible increase in performance of $0.7\%$ in terms of MODA when compared with our default approach using only foot regression. Removing all per-view losses resulted in a 0.3\% MODA decrease.

\begin{table}[!htbp]
\begin{center}
    \caption{BoosterSHOT (deformable transformer) performance by per-view loss}
    \label{table:per_view_loss}
    \begin{tabular}{c|ccccc}
    \toprule
    BoosterSHOT + Tr & MODA & MODP & precision & recall \\
    \midrule
    no loss   & 93.8 & 92.0 & 98.3 & 95.4 \\
    foot (default)    & 94.1 & 91.7 & 98.3 & 95.7 \\
    head + foot  & \textbf{94.8} & \textbf{92.1} & \textbf{98.4} & \textbf{96.3} \\
    \bottomrule
    \end{tabular}
\end{center}
\end{table}

\begin{figure}[!htbp]
    \centering
    \includegraphics[width=0.6\linewidth]{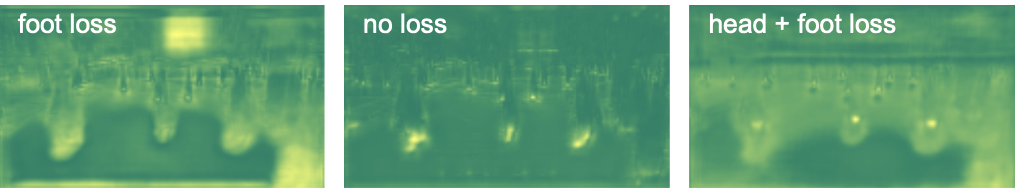}
    \caption{Comparison of image features with and without per-view loss}
	\label{fig:img_features_loss}
\end{figure}

In Figure~\ref{fig:img_features_loss}, we compare image features extracted from BoosterSHOT's feature extractor with different kinds of auxiliary loss settings. The features extracted when trained with no additional losses are more concentrated, while those from extractors trained with either foot regression loss or both foot and head regression losses are shown to be more broad. In addition, comparing the head + foot loss and foot loss feature heatmaps shows that the foot loss induces a bias toward the ground plane on which feet are placed, whereas the additional head loss counters that and helps the features attend overall to the entire body of the pedestrians. Table~\ref{table:per_view_loss} shows that including the head regression auxillary loss provides an additional increase in performance compared to only using the foot regression auxillary loss.

\subsection{Additional Results}
\label{app:F}
\begin{figure}[!htbp]
    \centering
    \label{fig:additional_channel_select_heatmap}
    \caption{Channel selection heatmap (Camera 2 to 6 from top to bottom)}
    \includegraphics[width=0.8\linewidth]{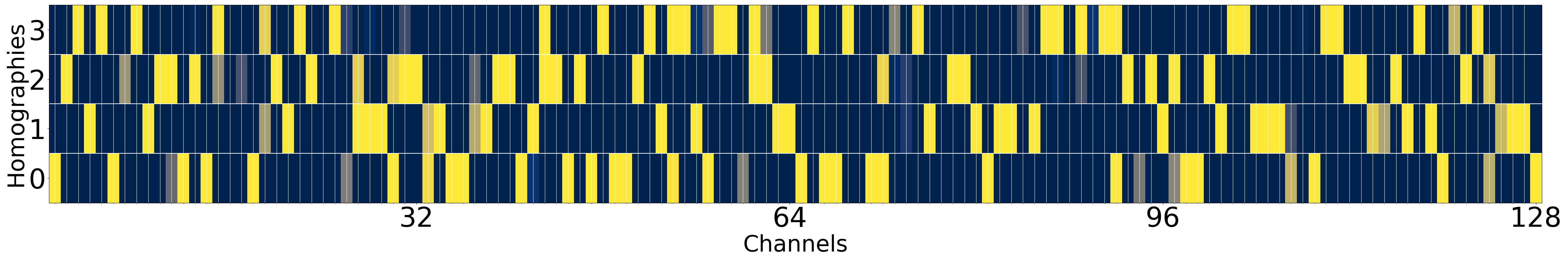}
    \includegraphics[width=0.8\linewidth]{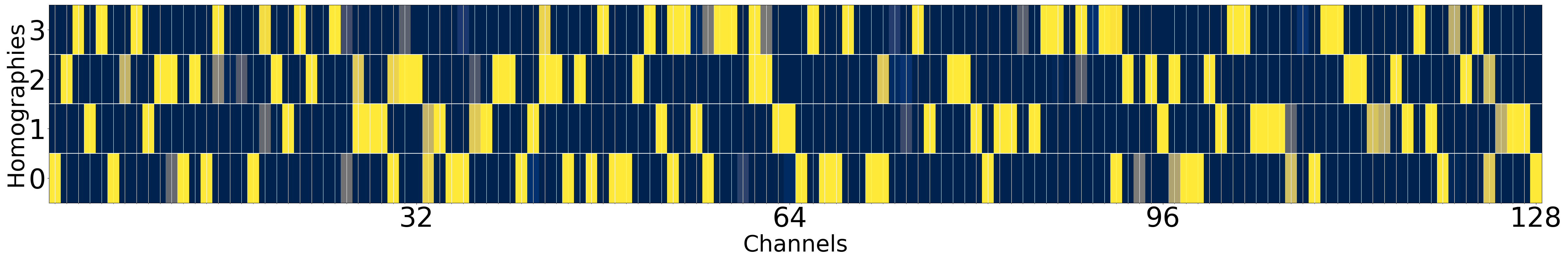}
    \includegraphics[width=0.8\linewidth]{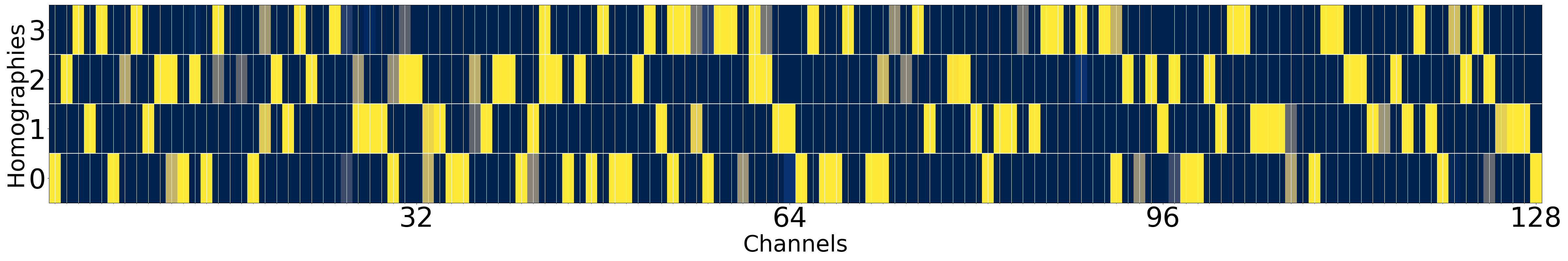}
    \includegraphics[width=0.8\linewidth]{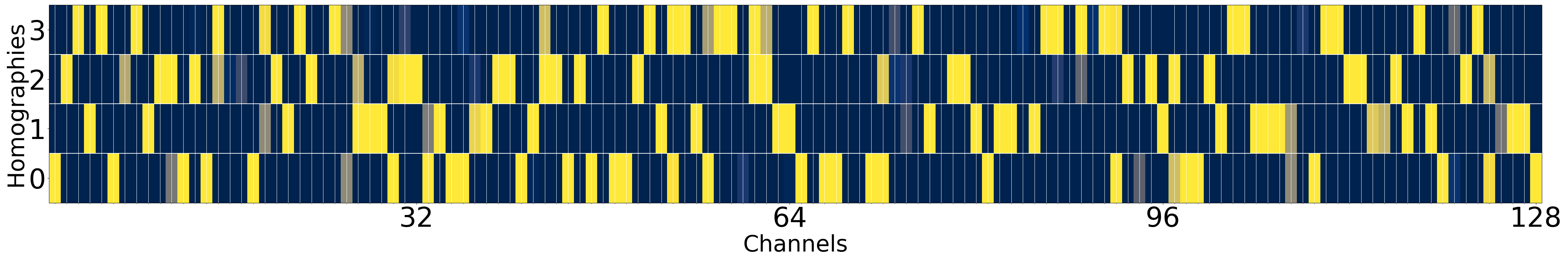}
    \includegraphics[width=0.8\linewidth]{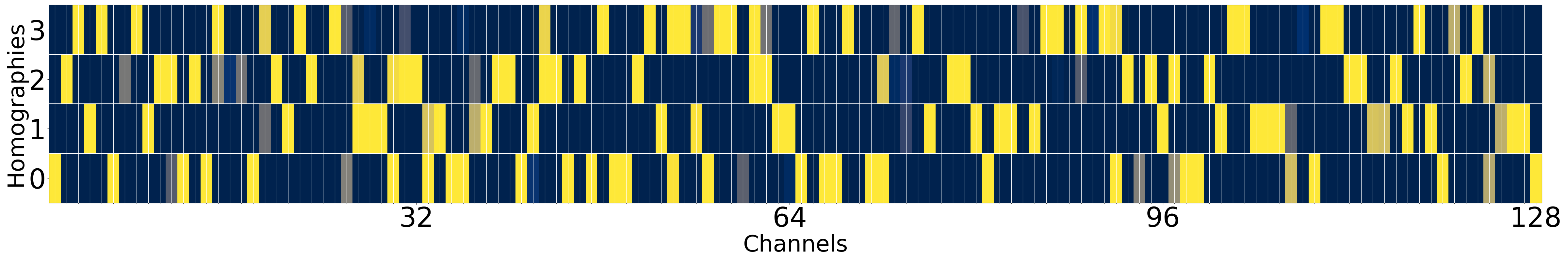}
\end{figure}

We include heatmaps showing which channels were selected for each homography in each camera view. Figure~\ref{fig:channel_select_heatmap} showed the selection heatmap for Camera 1 in MultiviewX, while Figure~\ref{fig:additional_channel_select_heatmap} shows the selection heatmaps for Cameras 2 through 6. Consistent with the previous findings, the channels that are selected for the majority of the test set for each homography are shown to be consistent across camera views.

\end{document}